\documentclass[letterpaper,10pt,journal,twoside]{IEEEtran}
\usepackage{ieee_robotics}
\usepackage{graphicx} \graphicspath{{figures/}}
\usepackage{amsmath,amssymb,mathtools,amsthm,nicefrac}
\usepackage[capitalise,noabbrev,nameinlink]{cleveref}
\usepackage[skip=3pt,font=small]{caption}
\usepackage{acronym}
\usepackage{wrapfig}
\usepackage[dvipsnames,svgnames,x11names,table]{xcolor}
\usepackage{enumitem}
\usepackage{xspace}
\usepackage[misc]{ifsym}
\usepackage{titletoc}
\usepackage{verbatim,fancyvrb,listings}
\usepackage{booktabs,tabularx,colortbl,multirow,array,makecell,tabularray}
\usepackage[detect-weight,mode=text]{siunitx}

\usepackage{orcidlink}
\usepackage[detect-weight,mode=text]{siunitx}
\crefname{ineq}{Ineq.}{Ineqs.}

\acrodef{imu}[IMU]{inertial measurement unit}
\acrodef{mse}[MSE]{mean squared error}
\acrodef{rmse}[RMSE]{root mean squared error}
\acrodef{sip}[SiP]{system-in-package}
\acrodef{fpc}[FPC]{flexible printed circuit}
\acrodef{mcu}[MCU]{microcontroller unit}
\acrodef{pcb}[PCB]{printed circuit board}
\acrodef{dof}[DoF]{degree of freedom}
\acrodefplural{dof}[DoFs]{degrees of freedom}
\acrodef{lbs}[LBS]{linear blend skinning}
\acrodef{stft}[STFT]{short-time Fourier transform}
\acrodef{spi}[SPI]{serial peripheral interface}
\newcommand{\name}{T-800\xspace}
\newcommand{\suppUrl}{https://player.vimeo.com/video/1148283536?h=42db8160dd}

\title{\name: An \SI{800}{\hertz} Data Glove\\for Precise Hand Gesture Tracking}
\author{Haoyang Luo\orcidlink{0009-0003-3342-7086}, Zihang Zhao\orcidlink{0000-0003-3215-7152}, Leiyao Cui\orcidlink{0009-0009-4925-6983}, Saiyao Zhang\orcidlink{0009-0007-2130-9110}, Liu Yang\orcidlink{0009-0000-3078-4940}, \\ Zhi Han\orcidlink{0000-0002-8039-6679},  Xiyuan Tang\orcidlink{0000-0003-2181-9042}, and Yixin Zhu\orcidlink{0000-0001-7024-1545}

\thanks{This work is supported in part by the Brain Science and Brain-like Intelligence Technology--National Science and Technology Major Project (2025ZD0219400), the National Natural Science Foundation of China (62376009), the State Key Lab of General AI at Peking University, the PKU-BingJi Joint Laboratory for Artificial Intelligence, the Wuhan Major Scientific and Technological Special Program (2025060902020304), the Hubei Embodied Intelligence Foundation Model Research and Development Program, and the National Comprehensive Experimental Base for Governance of Intelligent Society, Wuhan East Lake High-Tech Development Zone. (H. Luo and Z. Zhao contributed equally to this work.) ( Corresponding authors: Xiyuan Tang and Yixin Zhu.)}
\thanks{Haoyang Luo and Xiyuan Tang are with the Institute for Artificial Intelligence, Peking University, Beijing 100871, China, also with School of Integrated Circuits, Peking University, Beijing 100871, China (email: \texttt{xitang@pku.edu.cn}).}
\thanks{Zihang Zhao is with the Institute for Artificial Intelligence, Peking University, Beijing 100871, China, also with School of Psychological and Cognitive Sciences, Peking University, Beijing 100871, China, also with Beijing Key Laboratory of Behavior and Mental Health, Peking University, Beijing 100871, China, and also with the LeapZenith AI Research, Shanghai 201707, China.}%
\thanks{Leiyao Cui and Saiyao Zhang are with the State Key Laboratory of Robotics and Intelligent Systems, Shenyang Institute of Automation, Chinese Academy of Sciences, Shenyang 110016, China, also with the University of Chinese Academy of Sciences, Beijing 100049, China, and interned at the School of Psychological and Cognitive Sciences, Peking University, Beijing 100871, China.}%
\thanks{Liu Yang is with the School of Aerospace Engineering, Tsinghua University, Beijing 100084, China.}
\thanks{Zhi Han is with the State Key Laboratory of Robotics and Intelligent Systems, Shenyang Institute of Automation, Chinese Academy of Sciences, Shenyang 110016, China.}%
\thanks{Yixin Zhu is with the Institute for Artificial Intelligence, Peking University, Beijing 100871, China, also with School of Psychological and Cognitive Sciences, Peking University, Beijing 100871, China, and also with Beijing Key Laboratory of Behavior and Mental Health, Peking University, Beijing 100871, China (email: \texttt{yixin.zhu@pku.edu.cn}).}%
}

\begin{document}
\maketitle
\begin{abstract}
Human dexterity relies on rapid, sub-second motor adjustments, yet capturing these high-frequency dynamics remains an enduring challenge in biomechanics and robotics.
Existing motion capture paradigms are compromised by a trade-off between temporal resolution and visual occlusion, failing to record the fine-grained hand motion of fast, contact-rich manipulation.
Here we introduce \name, a high-bandwidth data glove system that achieves synchronized, full-hand motion tracking at \textbf{\SI{800}{\hertz}}.
By integrating a novel broadcast-based synchronization mechanism with a mechanical stress isolation architecture, our system maintains sub-frame temporal alignment across 18 distributed \acp{imu} during extended, vigorous movements.
We demonstrate that \name recovers fine-grained manipulation details previously lost to temporal undersampling. Our analysis reveals that human dexterity exhibits significantly high-frequency motion energy (\textbf{>\SI{100}{\hertz}}) that was fundamentally inaccessible due to the Nyquist sampling limit imposed by previous hardware constraints.
To validate the system's utility for robotic manipulation, we implement a kinematic retargeting algorithm that maps \name's high-fidelity human gestures onto dexterous robotic hand models. This demonstrates that the high-frequency motion data can be accurately translated while respecting the kinematic constraints of robotic hands, providing the rich behavioral data necessary for training robust control policies in the future.
\end{abstract}


\begin{figure*}[t!]
    \centering
    \includegraphics[width=\linewidth]{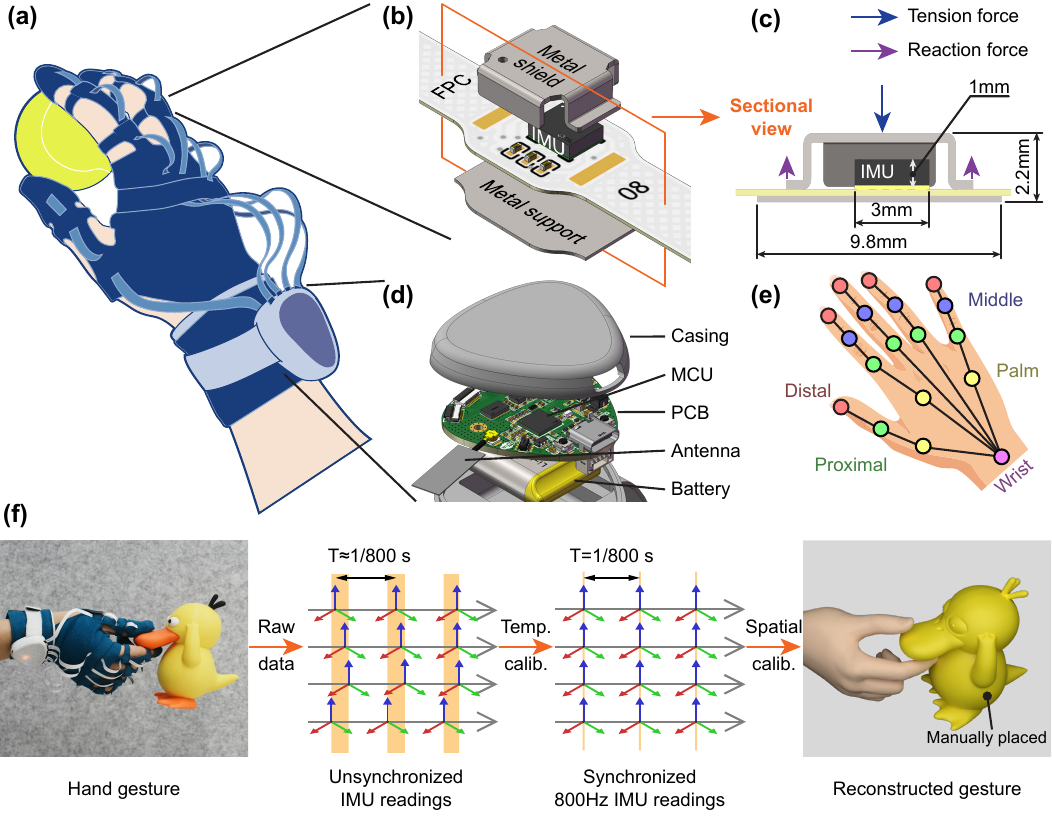}
    \caption{\textbf{System architecture and processing pipeline of the \name high-frequency tracking system.} (a) The wearable platform features an anatomically aligned, flexible sensor array designed to minimize restriction on natural hand movements. (b--c) \textbf{Mechanical isolation strategy.} (b) Exploded and (c) cross-sectional views of the ``sandwich'' structure. The rigid metal shield and support plate mechanically decouple the sensor from the fabric and skin; tension forces (blue arrows) and bone reaction forces (purple arrows) are routed through the rigid housing, ensuring stable coupling. (d) Wrist-mounted sensor hub consolidating heavy components (battery, \ac{mcu}) to minimize inertial loading on the finger segments. (e) Topology of the 18-sensor array covering the full kinematic chain. (f) \textbf{Spatiotemporal reconstruction workflow.} The pipeline transforms raw, asynchronous, and drifting data streams into a coherent \SI{800}{\hertz} signal via broadcast-based temporal calibration, followed by spatial alignment to reconstruct high-fidelity gestures.}
    \label{fig:overview}
\end{figure*}

\section{Introduction}

The human hand is a marvel of evolution, capable of executing sub-second, high-precision motor adjustments that remain the gold standard for artificial systems. Advances in high-\acs{dof} robotic hands have brought machines closer to this level of dexterity~\cite{deimel2016novel,shaw2023leap,zhao2025embedding,shadowrobot,Allegro}, enabling potential applications in minimally invasive surgery~\cite{vitiello2012emerging,li2024minitac}, precision assembly~\cite{li2022survey,keshvarparast2024collaborative}, unstructured household environments~\cite{zhao2025tac,zhao2025tacman,cui2025vi}, and \etc. However, the mechanical sophistication of these platforms introduces a control bottleneck: coordinating dozens of actuators in real-time to generate natural, agile behavior.

With the advancements of deep learning techniques, data-driven imitation learning has emerged as a promising solution by utilizing human demonstrations to guide robots through complex action spaces~\cite{qin2022dexmv,mandikal2022dexvip,arunachalam2023dexterous,jiang2025dexmimicgen,sheng2025learning}. Consequently, the fidelity of the recorded human motion datasets establishes an \textit{upper bound} on robotic performance~\cite{chen2025dataquality}. To reproduce the agility of a human hand, a system must capture the full spectrum of motion---particularly the high-frequency dynamics (>\SI{100}{\hertz}) that characterize rapid manipulation and contact responses~\cite{sundaram2019learning}.

Despite this critical need, current motion capture technologies face a fundamental dilemma between temporal resolution and spatial reliability. Optical systems (\eg, Vicon~\cite{vicon}) offer high sampling rates but suffer from severe occlusion during object interaction---precisely the moments when accurate tracking is most critical. Conversely, wearable \acf{imu}-based approaches allow for occlusion-free capture but are historically plagued by two limitations. First, sampling rates rarely exceed \SI{200}{\hertz}~\cite{connolly2017imu,edmonds2019tale,saggio2023dynamic,yang2021estimate,xue2025fi}, failing to satisfy the Nyquist sampling theorem for capturing rapid transient motions without aliasing. Second, independent sensor clocks drift over time, causing temporal misalignment that disperses simultaneous joint synergies across different timestamps. This ``temporal blurring'' degrades data integrity, rendering it unsuitable for learning precise dynamic models (\cref{tab:glove_comparison} provides a detailed comparison of representative state-of-the-art IMU-based hand motion capture systems).

To bridge this gap, we present \name, a high-frequency full-hand tracking system designed to capture the unseen dynamics of human dexterity. As illustrated in \cref{fig:overview}, \name integrates a compact 18-node \ac{imu} array with a reconfigurable communication protocol to achieve online synchronized \SI{800}{\hertz} acquisition. Unlike prior systems that rely on kinematic priors to hallucinate full-hand pose from fingertip \ac{imu} readings ~\cite{yang2021estimate,lu2023measurement,xue2025fi,lu2023online}, or from flex sensor data ~\cite{ahmed2021based, zou2024intelligent, fan2025data}, our system directly measures all finger segments. By quadrupling the bandwidth of conventional data gloves and enforcing sub-frame synchronization, \name provides a new opportunity to observe and analyze the high-speed spectral content of human manipulation.

\begin{table*}[t]
\renewcommand{\arraystretch}{1.3} 
\small
\centering
\begin{tabular}{@{}
                >{\raggedright\arraybackslash}m{1.8cm}
                >{\raggedright\arraybackslash}m{1.4cm}
                >{\raggedright\arraybackslash}m{1.2cm}
                >{\raggedright\arraybackslash}m{1.1cm}
                m{2.0cm}
                m{1.5cm}
                >{\centering\arraybackslash}m{1.5cm}
                >{\centering\arraybackslash}m{1.2cm}
                >{\centering\arraybackslash}m{1.0cm}
                >{\centering\arraybackslash}m{1.5cm}@{}}
\toprule
\multirow{2}{*}{Name} &
\multirow{2}{*}{\parbox{1.4cm}{\centering Sensor Type}} &
\multirow{2}{*}{\parbox{1.2cm}{\centering IMU Type}} &
\multirow{2}{*}{\parbox{1.1cm}{\centering Connectivity}} &
\multirow{2}{*}{\parbox{2.0cm}{\centering Haptic\\Obstruction}} &
\multirow{2}{*}{\parbox{1.5cm}{\centering Synchronization}} &
\multirow{2}{*}{\parbox{1.5cm}{\centering Palm \& Arm\\Sensors $\uparrow$}} &
\multirow{2}{*}{\parbox{1.2cm}{\centering Finger\\Sensors $\uparrow$}} &
\multirow{2}{*}{\parbox{1.0cm}{\centering IMU\\Count $\uparrow$}} &
\multirow{2}{*}{\parbox{1.5cm}{\centering Sample \\ Rate (Hz) $\uparrow$}} \\
\\
\midrule
iSEG-Glove~\cite{connolly2017imu} & IMU & MPU9150 & Wi-Fi & Fully-Obstructed & No & 1 & \textbf{3} & 16 & 40 \\
M. Edmonds \etal~\cite{edmonds2019tale} & IMU & BNO055 & Wired & Fully-Obstructed & No & 1 & \textbf{3} (2 for thumb) & 15 & 20 \\
G. Saggio \etal~\cite{saggio2023dynamic} & IMU & MPU9250 & Wired & Cutout Rings & No & \textbf{3} & \textbf{3} & \textbf{18} & 40 \\
Z. Yang \etal~\cite{yang2021estimate} & IMU + Magnetic & MPU9250 & Wired & Fingertip & Post-process & 1+0 & 1+1 & 3 & 100 \\
FI-HGR~\cite{xue2025fi} & IMU + Flex & ICM20948 & Bluetooth & Fully-Obstructed & Post-process & 1+0 & 1+0 (1+1 for index) & 6 & 50 \\
M.A. Ahmed \etal~\cite{ahmed2021based} & IMU + Flex & MPU9250 & Bluetooth & Fully-Obstructed & No & 1+1 & 1+1 & 6 & --- \\
C. Lu \etal~\cite{lu2023measurement} & IMU & ICM20948 & Wi-Fi & Finger Dorsal & Post-process & 2 & \textbf{3} & 5 & 200 \\
HIMT~\cite{zou2024intelligent} & IMU + Flex & BNO085 & Wi-Fi & Fully-Obstructed & Post-process & 2+0 & 0+1 & 2 & --- \\
ASTRA~\cite{bahrami2025astra} & IMU & MPU9250 & Wired & Fully-Obstructed & Post-process & 1 & \textbf{3} & 16 & 100 \\
L. Fan \etal~\cite{fan2025data} & IMU + Flex & MPU6050 & Bluetooth & Fully-Obstructed & No & 1+0 & 0+1 & 1 & --- \\
B. Lin \etal~\cite{lin2018design} & IMU & LSM9DSO & Bluetooth & Fully-Obstructed & No & 2 & \textbf{3} & 17 & 25 \\
C. Lu \etal~\cite{lu2023online} & IMU & ICM20648 & Bluetooth & Fully-Obstructed & No & 1 & 1 & 6 & 200 \\
Metagloves Pro~\cite{manusQuantumMeta} & IMU + Magnetic & --- & BLE & Palm \& Fingertip & --- & 1+0 & 1+1 & 6 & 120 \\
\midrule
\parbox{1.8cm}{\textbf{\name} \\ (This work)} & IMU & BHI360 & Wi-Fi & Cutout Rings & \textbf{Online} & \textbf{3} & \textbf{3} & \textbf{18} & \textbf{800} \\
\bottomrule
\end{tabular}
\caption{Comparison of different hand motion capture systems with IMU-based sensing. An upward arrow indicates that higher values are better.}
\label{tab:glove_comparison}
\end{table*}

Our primary contributions are:
\begin{itemize}[leftmargin=*,noitemsep,nolistsep]
    \item \textbf{High-Frequency Full-Hand Capture:} We develop a flexible glove architecture that achieves \SI{800}{\hertz} tracking across 18 joints. A novel mechanical isolation design (``sandwich structure'') ensures normal operation of the internal \ac{imu} while preserving natural hand dexterity.
    \item \textbf{Broadcast-Based Temporal Calibration:} We introduce a synchronization protocol that eliminates clock drift. By broadcasting a global timestamp latch command, we maintain sub-frame alignment across all distributed sensors throughout extended recording sessions.
    \item \textbf{Spectral Analysis of Dexterity:} Our system reveals that rapid hand-object interactions (\eg, pen spinning) contain substantial energy information above \SI{100}{\hertz}, proving that traditional \SI{200}{\hertz} systems undersample human agility.
    \item \textbf{Kinematic Retargeting}: We implement a geometric retargeting algorithm to map high-speed human hand data to dexterous robotic hand models, validating the kinematic compatibility of the capture data and laying the data foundation for future robust control algorithm training.
\end{itemize}

The remainder of this article is organized as follows. We begin by detailing the system architecture in \cref{sec:system_overview}, including the wearable platform design and mechanical isolation strategy. We then present our core algorithmic contributions: a broadcast-based synchronization protocol that eliminates clock drift (\cref{sec:temporal_calib}) and a spatial calibration framework for anatomical alignment (\cref{sec:spatial_calib}). In \cref{sec:high_speed_capture}, we validate the system's high-frequency capture capabilities through spectral analysis of rapid hand-object interactions. \cref{sec:retargeting} presents our kinematic retargeting framework and analyzes the mapping performance on robotic hand models. Finally, \cref{sec:conclusion} summarizes our findings and discusses future research directions, with comprehensive implementation details---including fabrication, hardware, and firmware specifications---provided in \cref{sec:methods}.

\begin{figure*}[t!]
    \centering
    \includegraphics[width=\linewidth]{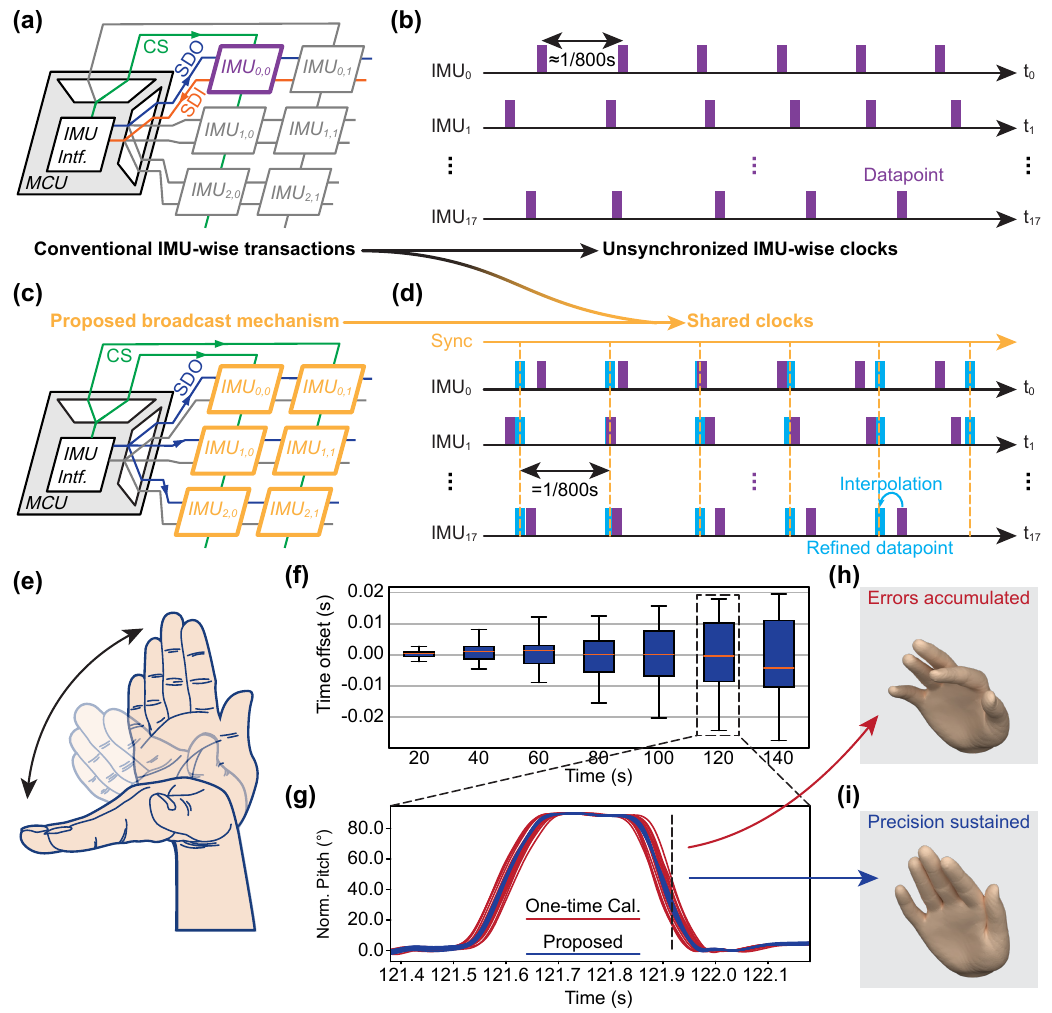}
    \caption{\textbf{Broadcast-based temporal calibration mechanism for drift-free multi-sensor synchronization.} (a--b) \textbf{The desynchronization bottleneck.} (a) Conventional architectures rely on sequential polling of individual nodes. (b) Each \ac{imu} runs on an independent internal oscillator. Slight frequency variations accumulate into significant clock drift, causing nominally simultaneous samples to scatter across the timeline (temporal jitter). (c--d) \textbf{Broadcast synchronization protocol.} (c) The proposed architecture utilizes a simultaneous broadcast command to latch a global timestamp across all \acsp{imu} instantly. (d) These latched timestamps serve as shared anchor points, enabling the system to interpolate asynchronous raw readings onto a unified, strictly aligned temporal grid. (e--i) \textbf{Validation via rapid dynamics.} (e) A rapid hand-flipping task (\SI{140}{\second} duration) serves as a stress test. (f) Quantitative analysis of time offsets reveals diverging clock drift across the 18 sensors using standard one-time calibration. (g) Signal coherence analysis: the proposed method (blue) maintains tight synchronization of finger pitch angles, whereas the baseline (red) exhibits severe temporal divergence. This divergence manifests visually as (h) the drifted hand gesture in the baseline reconstruction, contrasted with (i) the sustained kinematic fidelity and sub-frame precision achieved by our broadcast approach.}
    \label{fig:temporal_calib}
\end{figure*}

\section{Results and Discussion}\label{sec:sec2-result}
\subsection{\name System Overview}\label{sec:system_overview}

As illustrated in \cref{fig:overview}, the \name system is engineered to mitigate the trade-off in wearable motion capture: achieving high-fidelity, high-frequency kinematic tracking while imposing negligible physical restriction on natural hand dexterity. The system architecture integrates three tightly coupled subsystems: a mechanically isolated wearable sensing platform, a low-inertia wrist-mounted hub, and a real-time spatiotemporal reconstruction pipeline.

The wearable component (\cref{fig:overview}(a)) features a flexible glove substrate tailored to the complex anatomical curvature of the human hand (fabrication details in \cref{sec:methods-glove}). To ensure precise motion transduction, the system employs 18 distributed sensor nodes. A critical innovation in our design is the miniaturized \ac{imu} island, measuring just \SI{9.8}{\milli\metre} in length, \SI{6.8}{\milli\metre} in width, and \SI{2.2}{\milli\metre} in height. This ultra-compact form factor is specifically optimized to minimize inter-finger interference and proprioceptive obstruction during fine manipulation tasks. Furthermore, the island features a novel mechanical architecture designed to address the persistent challenge of firm sensor-to-bone coupling, as skin deformation and fabric sliding typically decouple sensors from underlying bone movement.

As depicted in the exploded and cross-sectional views of \cref{fig:overview}(b-c), each sensor node utilizes a ``sandwich'' structure. A custom metal top shield and a bottom support plate form a rigid exoskeleton that mechanically encapsulates the central \ac{imu}. This configuration creates a stress-shielding effect: tension forces exerted by the glove fabric (blue arrows) and reaction forces from the phalanges (purple arrows) are routed through the rigid outer housing, effectively bypassing the sensing element. Consequently, the \ac{imu} remains mechanically decoupled in the center, protected from shear stresses and pressure concentrations that would otherwise induce measurement errors. Simultaneously, this rigid interface concentrates the normal force to generate sufficient friction, ensuring firm, non-slip coupling between the sensor and the finger bone even during high-velocity movements.

The distributed sensor array is interconnected via a network of five custom-designed \acp{fpc} that conform to the hand's varying geometry. Unlike standard cabling, each \ac{fpc} segment is geometrically optimized for the specific range of motion of its corresponding joint---from the proximal knuckles to the distal fingertips. This design ensures that the interconnects possess sufficient compliance to accommodate full flexion without constraining joint mobility, yet remain taut enough to prevent excessive slack that could introduce sensor displacement. The specific material composition and layout of these flexible stripes are further detailed in \cref{sec:methods-stripe}.

To minimize the kinetic impact on hand dynamics, the \name design strategically consolidates all substantial mass components---including the \acf{mcu}, battery, wireless antenna, and main \acf{pcb}---into a single wrist-mounted sensor hub (\cref{fig:overview}(d)). By shifting the center of mass away from the distal segments to the wrist, this design substantially reduces the mass and inertia loaded onto the fingers, thereby preserving the user's natural agility during dynamic motions. The hub creates a protective enclosure for the electronics while serving as the central aggregation point for the distributed array. Hardware specifics of the sensor hub are provided in \cref{sec:methods-hub}.

The sensor hub interfaces with 17 \ac{imu} islands via the five \acp{fpc}, with an 18\textsuperscript{th} reference \ac{imu} embedded directly within the hub itself. The sensor topology is organized to align with the skeletal structure of the hand (\cref{fig:overview}(e)), with each \ac{fpc} branch hosting 3--4 nodes corresponding to the metacarpal and phalangeal bones of a single digit (color-coded from yellow to red). This comprehensive coverage ensures the capture of the complete kinematic chain from the wrist to the fingertips, leaving no joint unmeasured.

The end-to-end operation workflow is visualized in \cref{fig:overview}(f). The system operates as a distributed network where each \ac{imu} independently samples orientation data asynchronously at a nominal rate of \SI{800}{\hertz}. These raw, unsynchronized data streams, along with their local timestamps, are transmitted wirelessly to a host computer for real-time processing. To recover a coherent kinematic model from these independent streams, the host executes a two-stage calibration pipeline:

\paragraph*{Temporal Calibration}
Addressing the stochastic nature of independent sensor oscillators, this stage applies our broadcast-based synchronization algorithm (detailed in \cref{sec:temporal_calib}). It aligns the drifting local time bases of all 18 sensors to a unified global reference, effectively interpolating the irregular asynchronous samples into a strictly synchronized \SI{800}{\hertz} grid (\cref{fig:overview}(f)).

\paragraph*{Spatial Calibration}
To account for the inevitable physical variations in sensor placement, this stage computes the transformation between each \ac{imu}'s local frame and its corresponding anatomical bone frame using data collected during an initialization procedure (detailed in \cref{sec:spatial_calib}).

The final output is a high-fidelity, synchronized motion dataset where the orientation of every hand segment is expressed in a consistent world reference frame. This calibration pipeline facilitates the accurate reconstruction of complex interactions, such as the grasping demonstrated in \cref{fig:overview}(f).

\subsection{Multi-\texorpdfstring{\acs{imu}}{} Synchronization}\label{sec:temporal_calib}

Achieving precise temporal coherence across a distributed sensor array represents a challenge in wearable computing. In compact systems lacking dedicated synchronization lines, conventional architectures rely on the \ac{mcu} to sequentially poll individual nodes (\cref{fig:temporal_calib}(a)). This method suffers from a fundamental hardware limitation: each \ac{imu} operates on an independent internal oscillator. Due to manufacturing mismatches and temperature variations, these oscillators exhibit stochastic frequency deviations. This translates to a nominal offset from \SI{800}{\hertz} along with an unpredictable fluctuation in the sampling rate. These minute frequency discrepancies integrate over time, causing the local clocks to diverge gradually (\cref{fig:temporal_calib}(b)). Even with an initial ``one-time calibration,'' this accumulated drift inevitably results in millisecond-scale misalignment during extended recordings, destroying the temporal integrity of rapid, coordinated hand movements.

To eliminate this drift, the \name system implements a novel synchronization protocol built upon a reconfigurable communication architecture. Beyond standard one-to-one transactions, we programmed the \ac{mcu}'s I/O matrix to support a hardware-level broadcast mode (\cref{fig:temporal_calib}(c)). In this mode, a single ``capture-timestamp'' command is transmitted simultaneously onto the data lines and chip selection lines of all \acp{imu} through the I/O matrix. Upon command arrival, every sensor latches its current local timestamp. These globally synchronized events serve as absolute temporal anchors (\cref{fig:temporal_calib}(d)). The host computer utilizes these anchors to reconstruct a unified global timeline (implementation details in \cref{sec:methods-sync}). By modeling the relationship between each sensor's local clock and the master clock as a series of piecewise-linear segments constrained by successive broadcast anchors, the algorithm dynamically compensates for time-varying frequency drift. This allows the asynchronous raw data streams to be interpolated onto a strictly aligned \SI{800}{\hertz} grid with sub-frame precision.

The necessity of this approach is validated through a high-velocity hand-flipping experiment (\cref{fig:temporal_calib}(e)). \cref{fig:temporal_calib}(f) quantifies the temporal evolution of the 18 sensor clocks under standard one-time calibration; the time offsets exhibit divergence over a \SI{140}{\second} session. Such clock drift manifests as severe temporal divergence in the reconstructed motion. As shown in \cref{fig:temporal_calib}(h), the residual clock skew causes accumulated errors in the recorded motion, where the fingers appear to flip asynchronously despite being in a uniform motion. In contrast, our broadcast-based calibration maintains a tight, bounded error distribution throughout the session. The resulting reconstruction (\cref{fig:temporal_calib}(i)) preserves the temporal fidelity of the motion, ensuring that the kinematic phases of all fingers are aligned---a prerequisite for analyzing the high-frequency dynamics of dexterity.

\subsection{Full-Hand Gesture Reconstruction}\label{sec:spatial_calib}

\begin{figure}[t!]
    \centering
    \includegraphics[width=\linewidth]{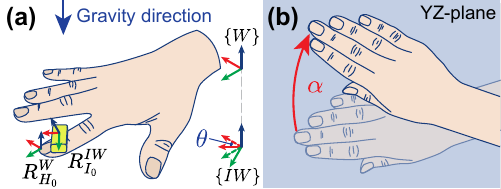}
    \caption{\textbf{Two-stage framework for calibrating sensor-to-bone mapping.} (a) Zero pose calibration: the hand is placed flat on a horizontal surface in the zero pose. The frames ${W}$ and ${IW}$ differ only by a rotation $\theta$ about their common $z$-axis (aligned antiparallel to gravity). (b) $\theta$ calibration: the hand is rotated by angle $\alpha$ about the world-frame $x$-axis against a vertical YZ-plane (\eg, a wall) to determine the unknown parameter $\theta$.}
    \label{fig:spatial_calibration}
\end{figure}

\begin{figure*}[t!]
    \centering
    \includegraphics[width=\linewidth]{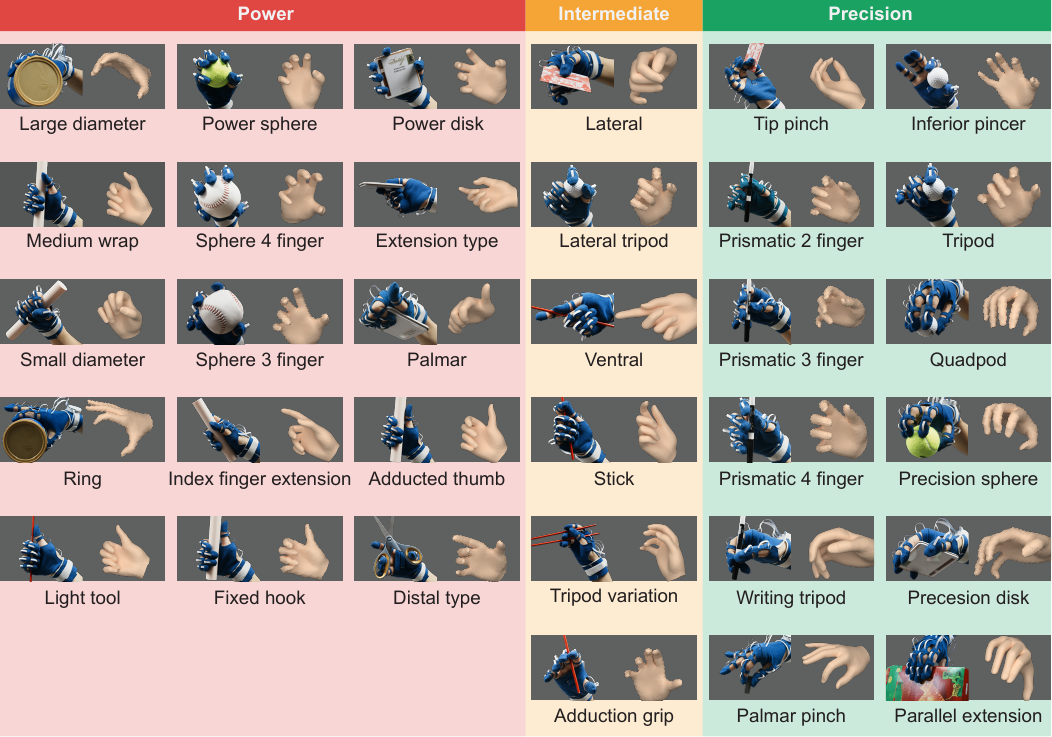}
    \caption{\textbf{Validation of kinematic fidelity across the 33-grasp taxonomy.} Systematic evaluation comparing physical hand configurations (left columns) with real-time reconstructions (right columns) captured by \name. The dataset is stratified according to the Feix \etal taxonomy~\cite{feix2015grasp} into three biomechanical categories: \textbf{Power grasps} (red) requiring high-force closure involving the palm; \textbf{Intermediate grasps} (yellow) balancing active manipulation with stability; and \textbf{Precision grasps} (green) demanding fine-grained fingertip articulation. The consistent reconstruction accuracy across this diverse range confirms two critical system properties: (i) the mechanical design preserves the hand's natural kinematic workspace without perceptible impedance; and (ii) the spatial calibration framework remains robust across the entire manifold of human hand poses.}
    \label{fig:static}
\end{figure*}

Transforming raw inertial measurements into a coherent biomechanical model requires the mapping from the sensor's arbitrary mounting frame to the anatomical joint frame. We derive a rigorous two-step framework to calibrate this mapping by leveraging gravity vectors and physical constraints.

\paragraph*{Notation and Frames}
We employ the notation ${R}^{j}_{i} \in SO(3)$ to denote the rotation matrix that transforms frame $\{i\}$ to frame $\{j\}$. Our calibration process involves four principal coordinate frames, as illustrated in \cref{fig:spatial_calibration}: the world frame $\{W\}$, the \ac{imu} world frame $\{IW\}$, the hand frame $\{H\}$, and the \ac{imu} body frame $\{I\}$.

The \ac{imu} world frame $\{IW\}$ is established at sensor initialization and remains fixed thereafter. A key property of $\{IW\}$ is that its $z$-axis aligns antiparallel to the gravity vector, as determined during initialization. The rotation ${R}^{IW}_{I}$ represents the current \ac{imu} orientation measurement with respect to this initial frame. Without loss of generality, we define the world frame $\{W\}$ such that its $z$-axis also opposes gravity. Under this definition, frames $\{W\}$ and $\{IW\}$ differ only by a rotation $\theta$ about their common $z$-axis, yielding the rotation matrix
\begin{equation}
{R}^{W}_{IW} =
\begin{bmatrix}
\cos{\theta} & -\sin{\theta} & 0\\
\sin{\theta} & \cos{\theta} & 0\\
0 & 0 & 1
\end{bmatrix}.
\label{eq:world_align}
\end{equation}

\paragraph*{Problem Formulation}
Our reconstruction objective is to recover the anatomical hand orientation ${R}^{W}_{H}$ given the instantaneous sensor measurement ${R}^{IW}_{I}$. The kinematic chain is governed by the general transformation:
\begin{equation}
    {R}^{W}_{H} = {R}^{W}_{IW}{R}^{IW}_{I}{R}^{I}_{H}.
    \label{eq:general_form}
\end{equation}
This equation contains two unknown static transformations: the global heading offset ${R}^{W}_{IW}$ (parameterized by $\theta$) and the bone-to-sensor mounting orientation ${R}^{I}_{H}$. We solve for these unknowns via a constraint-based calibration protocol.

\paragraph*{Step 1: Zero Pose Calibration}
As shown in \cref{fig:spatial_calibration}(a), we first place the hand flat on a horizontal surface, such as a tabletop, with the palm facing downward. We define the forward direction of the hand as the $y$-axis of the world frame and designate this configuration as the zero pose:
\begin{equation}
{R}^{W}_{H_0} = I,
\label{eq:init_pose}
\end{equation}
where $I$ denotes the $3 \times 3$ identity matrix. Substituting \cref{eq:init_pose} into \cref{eq:general_form} and recording the corresponding \ac{imu} reading ${R}^{IW}_{I_0}$, we can express ${R}^{I}_{H}$ in terms of ${R}^{W}_{IW}$:
\begin{equation}
{R}^{I}_{H} = {R}^{I}_{IW}{R}^{IW}_{W}R^{W}_{H}= {R}^{I_0}_{IW}{R}^{IW}_{W}.
\end{equation}
This substitution leverages the more tractable form of ${R}^{W}_{IW}$ established in \cref{eq:world_align}, reducing our unknowns to the single parameter $\theta$.

\paragraph*{Step 2: $\theta$ Calibration}
To determine $\theta$, we rotate the hand about the world-frame $x$-axis by an angle $\alpha \in (0, \pi)$, as illustrated in \cref{fig:spatial_calibration}(b). This rotation can be performed by placing the hand against a vertical YZ-plane, such as a wall. We record the \ac{imu} readings at the start and end of this rotation, denoted by ${R}^{IW}_{I_s}$ and ${R}^{IW}_{I_e}$, respectively. Leveraging \cref{eq:general_form}, the relationship between these frames can be expressed as
\begin{equation}
R^{IW}_{I_0}R^{Is}_{IW}R^{IW}_{I_e}R^{I_0}_{IW} = R^{IW}_{W}R^{H_s}_{H_e}R^{W}_{IW},
\end{equation}
where $R^{H_s}_{H_e}$ represents the rotation by angle $\alpha$ about the $x$-axis. For notational simplicity, we denote the known left-hand side as $R$. Expanding the right-hand side yields
\begin{equation}
\resizebox{0.9\columnwidth}{!}{%
$\begin{aligned}
R = &
\begin{bmatrix}
\cos{\theta} & \sin{\theta} & 0\\
-\sin{\theta} & \cos{\theta} & 0\\
0 & 0 & 1
\end{bmatrix}
\begin{bmatrix}
1 & 0 & 0\\
0 & \cos{\alpha} & -\sin{\alpha} \\
0 & \sin{\alpha} & \cos{\alpha}
\end{bmatrix}
\begin{bmatrix}
\cos{\theta} & -\sin{\theta} & 0\\
\sin{\theta} & \cos{\theta} & 0\\
0 & 0 & 1
\end{bmatrix}\\
= &
\begin{bmatrix}
* & * & -\sin{\theta} \sin{\alpha}\\
* & * & -\cos{\theta} \sin{\alpha}\\
* & * & *
\end{bmatrix},
\end{aligned}$%
}
\end{equation}
where asterisks denote elements not needed for our solution. Since $\alpha \in (0, \pi)$, we have $\sin{\alpha}>0$. By examining the third column, we can directly solve for $\theta$:
\begin{equation}
\theta = \text{atan2}(-R_{13}, -R_{23}),
\end{equation}
where $R_{ij}$ denotes the element of $R$ in the $i$-th row and $j$-th column. With $\theta$ determined, both unknown transformations ${R}^{W}_{IW}$ and ${R}^{I}_{H}$ are fully specified, completing the calibration procedure.

\paragraph*{Validation}
This closed-form solution allows us to simultaneously compute the 18 unique sets of calibration matrices for the entire hand in a single pass. To validate the kinematic fidelity of this approach, we evaluated the system against the standard 33-grasp taxonomy~\cite{feix2015grasp}. As visualized in \cref{fig:static}, \name accurately reconstructs the full spectrum of human grasp types---ranging from power grasps to precision pinches---without kinematic singularities, confirming that the spatial calibration effectively factorizes out sensor mounting heterogeneity.

\begin{figure*}[ht!]
    \centering
    \includegraphics[width=\linewidth]{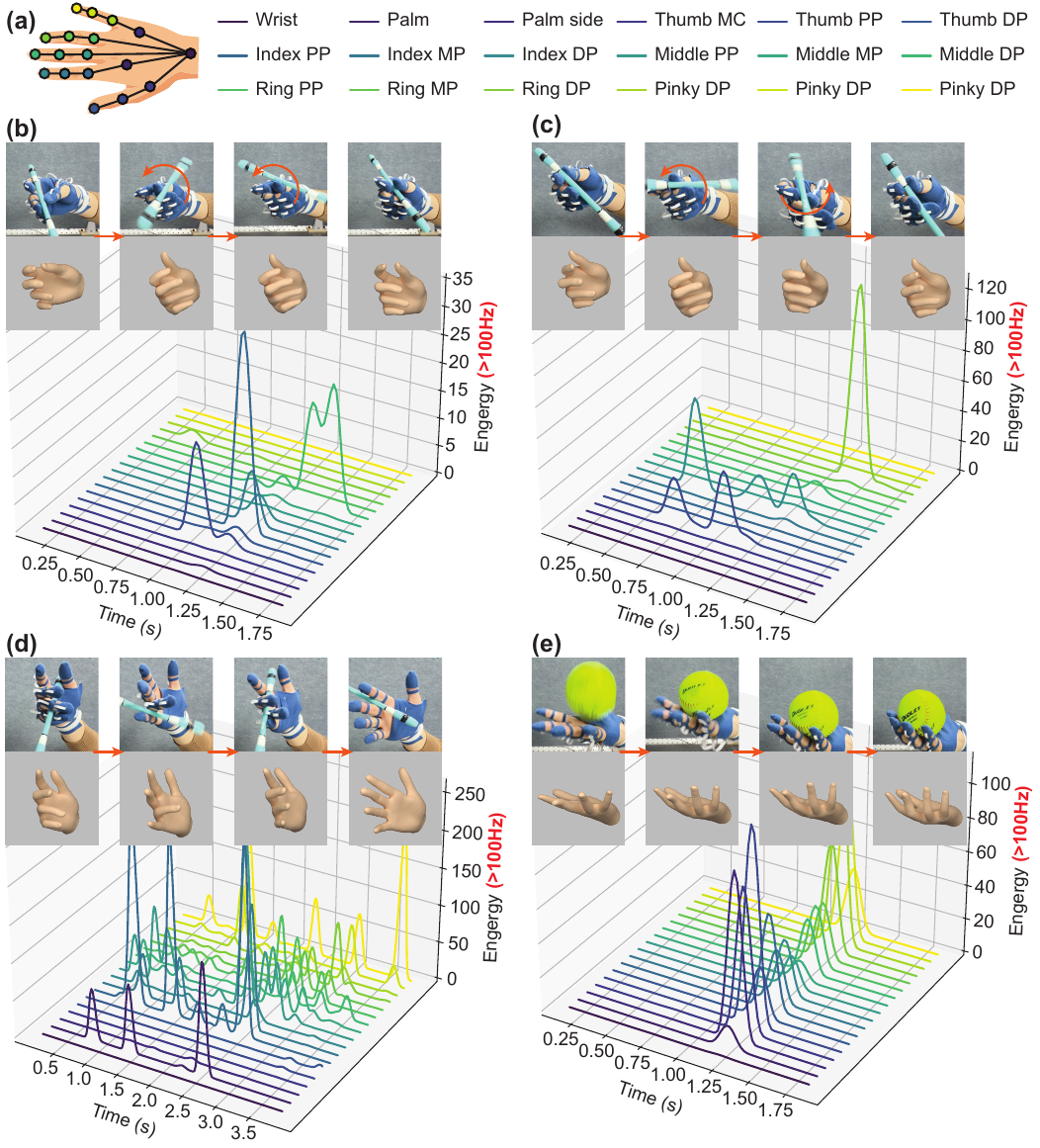}
    \caption{\textbf{Spectral discovery of high-frequency kinematics in human dexterity.} (a) Anatomical sensor mapping of the 18-node array (MC: metacarpal, PP: proximal, MP: middle, DP: distal phalanx). (b--e) \textbf{Biomechanical energy landscapes.} Time-resolved spectral power summation for frequencies strictly $>\SI{100}{\hertz}$---a band theoretically invisible to conventional \SI{200}{\hertz} systems. (b--c) Pen spinning tasks (unidirectional and bidirectional) reveal distinct, high-amplitude spectral bursts spatially localized to the manipulating digits (Thumb, Index, Middle). These transient spikes coincide with rapid actuation phases, confirming that fine motor control relies on sub-millisecond micro-adjustments. (d) In a four-finger spinning task explicitly excluding the thumb, the thumb sensor exhibits spectral quiescence (flatline). (e) Catching a heavy object elicits a synchronized, global high-frequency response across the entire hand. Together, these profiles validate that signal energy $>\SI{100}{\hertz}$ is kinematic information essential for high-fidelity modeling. Detailed manipulation processes and reconstructed gestures are available in the \href{\suppUrl}{Supplementary Video}.}
    \label{fig:dynamic}
\end{figure*}

\subsection{Spectral Analysis of High-Speed Manipulation}\label{sec:high_speed_capture}

The 18-channel synchronized acquisition capability of \name, operating at a stable \SI{800}{\hertz}, provides a unique observational window into the spectral composition of human dexterity. By extending the bandwidth limitations of prior hardware, we can now rigorously investigate the high-frequency kinematic information ($> \SI{100}{\hertz}$) that may exist in the rapid manipulation tasks. To probe this, we subjected the system to a set of dynamic tasks involving fine motor control (pen spinning) and impact absorption (tossing and catching), as visualized in \cref{fig:dynamic}.

\paragraph*{Frequency Domain Signal Processing}
We quantify the spectral power density of hand motion through a systematic signal processing pipeline designed to extract transient kinematic features. First, we compute the instantaneous magnitude of the angular velocity vector for each anatomical segment. To localize transient spectral features in the time domain, we apply a \acf{stft} to the angular velocity signal. We utilize a Hann window of length 256 samples combined with a sliding step size of 16 samples, yielding time-varying energy strength at each frequency point. 

Crucially, our analysis isolates spectral energy components strictly exceeding the \SI{100}{\hertz} threshold. This cutoff targets the blind spot of current technology. Since conventional wearable \ac{imu} systems typically operate at sampling rates below \SI{200}{\hertz}, their signal reconstruction capacity is mathematically hard-capped at \SI{100}{\hertz} by the Nyquist sampling theorem. Consequently, any energy detected by \name above this threshold represents biomechanical dynamics that have remained fundamentally invisible to previous generations of data gloves. Therefore, the detection of spectral content in this band serves as definitive proof of the necessity for higher sampling rates.

\paragraph*{Biomechanical Energy Profiles}
The resulting energy landscapes, plotted as the linear summation of high-frequency components ($> \SI{100}{\hertz}$) (without logarithmic scaling to emphasize differences) in \cref{fig:dynamic}(b)--(e), reveal that human dexterity exhibits rich spectral content well beyond the sampling limits of previous hardware.

In the precision manipulation task involving counterclockwise pen spinning (\cref{fig:dynamic}(b)), which recruits the thumb, index, and middle fingers, we observe sharp, distinct bursts of high-frequency energy. These bursts are spatially localized exclusively to the three manipulating digits, appearing in temporal synchrony with the rapid actuation phases required to maintain object rotation. Extending this to a bidirectional task (\cref{fig:dynamic}(c)) introduces additional high-frequency signatures, corresponding to the complex rapid deceleration and re-acceleration required to reverse the pen's angular momentum.

We further showed a four-finger spinning task that explicitly excludes the thumb (\cref{fig:dynamic}(d)). The resulting energy profile demonstrates intense spectral activation across the index, middle, ring, and pinky fingers, as well as the palm. In stark contrast, the thumb and forearm segments remain spectrally quiescent (exhibiting a flat energy response). This negative control confirms that the recorded high-frequency signals are genuine kinematic features of the moving segments, rather than sensor noise or mechanical resonance propagated globally through the glove structure.

Finally, in the heavy softball tossing task (\cref{fig:dynamic}(e)), the system captures the phase of impact absorption. Unlike the localized finger dynamics of pen spinning, this task elicits a synchronized, high-energy response across the entire hand (excluding the lower arm) as the fingers and palm work in unison to absorb the ball's impact.

\paragraph*{Insights into Dexterous Manipulation}
These results collectively demonstrate two key insights. First, dexterous hand motions contain rich high-frequency components above \SI{100}{\hertz} that are beyond the reach of existing motion capture systems. Second, \name's high sampling rate and synchronized multi-sensor design enable precise localization of these high-frequency components to specific fingers and hand segments involved in the manipulation. This selective activation pattern validates both the spatial accuracy of our calibration and the temporal fidelity of our capture system, opening new possibilities for understanding and analyzing complex hand motions.

\begin{figure*}[ht!]
    \centering
    \includegraphics[width=\linewidth]{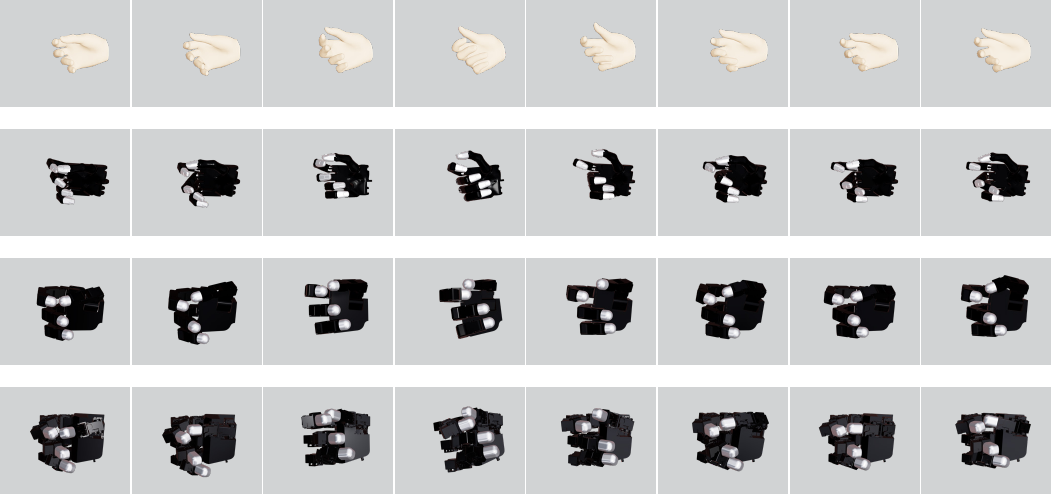}
    \caption{\textbf{Kinematic retargeting validation across diverse robotic hand platforms.} Representative frames from dynamic manipulation tasks demonstrating the mapping of captured human hand motions by \name (top row) to three robotic hand models with progressively diverse kinematic structures: Shadow Dexterous Hand, Allegro Hand, and Leap Hand. The sequential frames illustrate the algorithm's ability to transfer high-frequency human manipulation dynamics to robotic platforms while respecting their kinematic constraints.}
    \label{fig:retarget}
\end{figure*}

\subsection{Kinematic Retargeting to Robotic Hands}\label{sec:retargeting}

To validate the kinematic compatibility of the captured high-frequency data with robotic systems, we devise a retargeting algorithm that maps human hand motion to a dexterous robotic hand model. The objective is to compute the optimal robot joint configuration $\mathbf{q} \in \mathbb{R}^d$ that minimizes the Euclidean distance between key kinematic features of the human and robotic hands, relative to their respective wrist frames.

\paragraph*{Problem Formulation}
We formulate this as a constrained optimization problem. For each finger $f$, we designate the objective as the fingertip position $\mathbf{p}_{t}$ to ensure manipulation precision. Let $\mathbf{p}^{r}$ denote the target position derived from the \name glove data and $\mathbf{p}(\mathbf{q})$ denote the position computed via the robot's forward kinematics. To solve for the joint updates, we treat the instantaneous tracking error as a vector $\mathbf{e} \in \mathfrak{se}(3)$ within the Lie algebra of the rigid body transformation group. The retargeting problem is formulated as:
\begin{equation}
\begin{aligned}
\min_{\mathbf{q}} &\quad\sum_{f=1}^{N_f} \| \Lambda
\mathbf{e}_{t,f}(\mathbf{q})\|_2 \\
\text{s.t.} &\quad\mathbf{q}_{l} \preccurlyeq \mathbf{q} \preccurlyeq \mathbf{q}_{u}
\end{aligned},
\end{equation}
where $N_f$ is the number of fingers, and $\Lambda \in \mathbb{R}^{6\times6}$ is a diagonal selection matrix with the first three elements set to one and the remainder to zero, effectively isolating the translation component of the twist. The terms $\mathbf{q}_{l}$ and $\mathbf{q}_{u}$ represent the lower and upper joint limits of the robotic hand, respectively.

\paragraph*{Sequential Programming}
To solve this nonlinear optimization efficiently, we linearize the forward kinematics around the current configuration $\mathbf{q}$ via the first-order approximation $\Delta \mathbf{p} \approx \mathbf{J}_e \Delta \mathbf{q}$, and apply a sequential programming framework~\cite{zhao2025b,nocedal1999numerical}. This approach iteratively refines the solution while adaptively adjusting the step size based on the linear model's approximation quality and the satisfaction of joint constraints.

We derive the effective task Jacobian $\mathbf{J}_e$ via a chain rule that relates changes in joint space to changes in the Lie algebra error vector. This composition involves the standard frame Jacobian $\mathbf{J}(\mathbf{q})$, which maps joint velocities to the end-effector frame pose $E\in\mathrm{SE(3)}$, and the right derivative $\mathbf{J}(E)$ of the logarithmic map, which relates the pose to the Lie algebra error coordinate:
\begin{equation}
\mathbf{J}_{e} = \Lambda \mathbf{J}(E) \mathbf{J}(\mathbf{q}).
\end{equation}

\paragraph*{Validation}
To verify the versatility of the proposed framework and the kinematic compatibility of the \name data, we evaluated the retargeting algorithm on three distinct robotic hands with progressively diverse kinematic structures: the five-finger Shadow Dexterous Hand~\cite{shadowrobot} (24 \acsp{dof}), the four-finger Allegro Hand~\cite{Allegro} (16 \acsp{dof}), and the four-finger Leap Hand~\cite{shaw2023leap} (16 \acsp{dof}). This selection spans a spectrum from highly anthropomorphic designs to more constrained, lower-\acp{dof} architectures, rigorously testing the algorithm's ability to accommodate substantial kinematic disparities.

We applied the retargeting algorithm to dynamic human manipulation tasks captured in \cref{sec:high_speed_capture}, with representative frames illustrated in \cref{fig:retarget}. Performance was quantified using \ac{rmse}:
\begin{equation}
\mathrm{RSME} = \sqrt{\frac{1}{3N_f}\sum_{f=1}^{N_f} \left\| \Lambda \mathbf{e}_{t,f}(\mathbf{q})\right\|_2^2}.
\end{equation}

The results demonstrate precise motion transfer across all platforms, with performance closely correlating to kinematic capabilities. Paradoxically, the Shadow Hand, despite having the closest anatomical correspondence to the human hand, achieved the highest \ac{rmse} (\SI{8.3 \pm 6.8}{\milli\meter}). This is primarily because its compact design constrains the range of motion for each \acs{dof}. In contrast, the Allegro Hand and Leap Hand achieved competitive performance (\SI{1.0 \pm 1.8}{\milli\meter}, and \SI{0.6 \pm 1.4}{\milli\meter}, respectively), despite their four-finger configuration introducing significant kinematic mismatch. Their superior performance stems from two key factors: larger \acs{dof} ranges and the optimizer's ability to exploit redundant joint null spaces to compensate for the anatomical differences.

These results confirm that the captured data is not only biomechanically accurate but also directly transferable for controlling diverse robotic platforms, validating both the data quality and the framework's robustness to kinematic variations.

\section{Conclusion and Outlook}\label{sec:conclusion}

In this work, we have bridged the gap between high-speed kinematic capture and unhindered human dexterity with \name, a rapid full-hand tracking system that extends the bandwidth limitations of prior \ac{imu}-based systems. By fusing a stress-shielding mechanical architecture with a broadcast-based synchronization protocol, we achieved continuous, drift-free motion tracking at \SI{800}{\hertz} across 18 anatomical joints. Crucially, this system allowed us to empirically uncover the spectral richness of human manipulation, revealing that significant biomechanical energy exists beyond \SI{100}{\hertz}---the Nyquist limit of previous hardware. These high-frequency dynamics---previously masked by frequency aliasing---are shown to be carrying valuable information in fine motor tasks like precision spinning and impact absorption.

The implications of \name extend beyond biomechanical analysis to become an enabling infrastructure for embodied intelligence. By providing a source of occlusion-free, high-fidelity human demonstrations, this platform addresses the data quality bottleneck in robotic imitation learning. Algorithms trained on this ``super-resolution'' kinematic data can potentially acquire agile behaviors that were previously unlearnable due to signal smoothing~\cite{qin2022dexmv,mandikal2022dexvip,arunachalam2023dexterous,jiang2025dexmimicgen,li2026simultaneous}.

Looking forward, the natural evolution of this platform lies in the convergence of kinematics and kinetics. Integrating dense tactile sensing arrays into the glove structure presents a compelling frontier: coupling spatiotemporal contact maps with our synchronized high-speed motion data would capture the complete sensorimotor loop of human manipulation. Such multimodal datasets are essential for learning bidirectional mappings between tactile feedback and motor control, grounding robotic policies in the physics of contact~\cite{johansson2009coding,jiang2024triboelectric,li2025taccel,an2026flexible}. Ultimately, through open-source hardware benchmarks and standardized data formats, we envision \name serving as a foundational tool to catalyze the community's transition from quasi-static grasping to truly dynamic, human-level dexterity.

\section{Methods}\label{sec:methods}

\subsection{Fabrication of the Wearable Platform}\label{sec:methods-glove}

The wearable substrate is constructed from high-elasticity polyester fabrics, selected for their durability and conformability. The architecture features a dorsal backbone tailored to the hand's topology, extended by a palmar strap that anchors the two palm sensors. To ensure robust sensor-to-bone coupling without adhesives, we implemented a tension-based fixturing mechanism. Elastic rings (positioned at the proximal and intermediate phalanges) and fingertip caps are manufactured to approximately \SI{90}{\percent} of the average human's corresponding finger diameter. This intentional undersizing induces a controlled pre-tension force upon donning.

The \ac{imu} islands are seated beneath the overlapping sections of the dorsal backbone. The normal force generated by the pre-tensioned elastic components creates sufficient friction to mechanically lock the islands in place, minimizing relative sliding during vigorous dynamic movements.

\subsection{Flexible Interconnects and Sensor Integration}\label{sec:methods-stripe}

The distributed sensor array is interconnected via a custom multi-layer \acp{fpc} designed to withstand repeated bending cycles. The stack-up comprises two \SI{35}{\micro\meter} copper foil layers encapsulated between three layers of \SI{25}{\micro\meter} polyimide, bonded by two additional \SI{25}{\micro\meter} adhesive interlayers. The geometric profile of each \ac{fpc} stripe is parametrically optimized to match the outer contour length of the corresponding digit at maximum flexion (fist gesture), preventing constraining hand movements.

Each sensor node integrates a 6-axis \ac{imu} (BHI360, Bosch Sensortec) featuring triaxial gyroscopes and accelerometers. To implement the stress-shielding ``sandwich'' architecture, the \ac{fpc} is mechanically reinforced by a steel support plate (bottom) and a protective steel shield (top). Manufacturing of the ``sandwich'' involves a two-stage process: the bottom support is hot-pressed onto the \ac{fpc} substrate for structural rigidity, while the top shield is soldered post-assembly.

Prior to deployment, each sensor unit undergoes a calibration routine utilizing static measurements. This process compensates for deterministic error sources, including gyroscope bias drift, accelerometer zero-g offset, and cross-axis sensitivity matrices. The resulting calibration parameters are stored in the sensor hub's non-volatile memory and uploaded to the \acp{imu} during system initialization.

\subsection{Sensor Hub Electronics}\label{sec:methods-hub}

The wrist-mounted sensor hub (\cref{fig:overview}(d)) is architected around a highly-integrated wireless \ac{sip} (ESP32-S3-PICO-1, Espressif Systems). It integrates dual-core 32-bit microprocessors, a \SI{2.4}{GHz} Wi-Fi-capable radio, and a configurable I/O matrix. To interface the \ac{mcu} I/O voltage (\SI{3.3}{V}) with the low-power sensor array (\SI{1.8}{V}), the custom \ac{pcb} incorporates dedicated level-shifting logic. The system is powered by a compact lithium-ion polymer battery (Model 802025, \SI{400}{mAh}) and regulated by a high-efficiency power management integrated circuit (ETA6093), which yields an operating time of approximately 2 hours.

Data transactions between the \ac{sip} and the \acp{imu} are executed via high-speed \ac{spi} buses. To minimize wire count, sensors on the same finger stripe share common clock and data lines, with individual addressing managed via dedicated chip-select lines.

\subsection{Firmware Architecture for Synchronized Acquisition}\label{sec:methods-sync}

Upon power-up, each \ac{imu} is configured to sample local orientation at a fixed \SI{800}{\hertz}, timed by its internal oscillator. To achieve a stable end-to-end latency, the firmware employs a deterministic superloop architecture that decouples time-critical signaling from data processing.

\paragraph*{Command Broadcast}
To implement the global synchronization mechanism, we modified the \ac{mcu}'s hardware abstraction layer to operate the I/O matrix in broadcast mode. This allows a single internal signal to be routed simultaneously to multiple physical lines. This technique is applied to \ac{spi} clock lines, data output lines, and chip-select lines during the broadcast (\cref{fig:temporal_calib}(c)), ensuring sub-frame synchronization of the timestamp latch command.

\paragraph*{Data Retrieval}
Following the broadcast latch, the firmware initiates sequential data retrieval. To maximize throughput, the two independent hardware \ac{spi} controllers are utilized in parallel, allowing simultaneous polling of two sensor stripes.

\paragraph*{Core Affinity}
To eliminate time jitter from RTOS scheduling in the \ac{sip}, the task-critical superloop for broadcasting and SPI transaction is pinned to one microprocessor core (Core 0), while the second core (Core 1) handles the Wi-Fi protocol stack and background housekeeping. This separation ensures that the \SI{800}{\hertz} acquisition cycle remains strictly periodic and uninterrupted by wireless transmission overhead.

\subsection{Host-Side Telemetry and Real-Time Reconstruction}\label{sec:methods-host}

The host computer functions as the centralized aggregation node for kinematic synthesis. To preserve signal integrity, raw orientation packets and hardware timestamps are streamed from the sensor hub using a hybrid network topology: while the hub transmits wirelessly via Wi-Fi, the host workstation is tethered to the router via an Ethernet cable. This configuration minimizes wireless contention and packet latency jitter. At the protocol level, sensor data is encapsulated into fixed-length payloads to ensure deterministic transmission latency and prevent fragmentation overhead.

Upon receiving, the binary stream is unpacked into \ac{imu}-specific floating-point quaternions and timestamp tuples. A temporal calibration engine utilizes the anchor points captured during the broadcast synchronization phase to map each sensor's local clock to the global master clock. This mapping enables two critical operations: (i) \textit{resampling}, where asynchronous measurements are interpolated onto a uniform, time-aligned grid; and (ii) \textit{drift monitoring}, which continuously tracks oscillator deviation to verify synchronization stability independent of the calibration algorithm.

Following temporal alignment, the data undergoes spatial transformation. The calibration matrices derived in \cref{sec:spatial_calib} (${R}^{W}_{IW}$ and ${R}^{I}_{H}$) are applied to the raw readings, mapping local sensor frames to anatomical bone frames in world coordinates. This synchronized, physically registered kinematic stream constitutes the primary dataset, logged for offline analysis.

\section*{Acknowledgments}
Our sincere thanks go to Miss Qing Gao (Peking University) and Mr. Haikang Diao (Peking University) for data collection, Prof. Wei Liang (Beijing Institute of Technology) and Prof. Bing Ning (Beijing Institute of Fashion Technology) for their help in glove manufacturing, Mr. Lei Yan (LeapZenith AI Research) for mechanical design support, and Ms. Hailu Yang (Peking University) for her assistance in procuring every piece of raw material necessary for this research.

\bibliography{reference_header,reference}
\bibliographystyle{IEEEtran}

\end{document}